\documentclass{article}

 \usepackage[dvipsnames]{xcolor}

      \usepackage[preprint]{neurips_2020}

\usepackage[utf8]{inputenc} 
\usepackage[T1]{fontenc}    
\usepackage{hyperref}       
\usepackage{url}            
\usepackage{booktabs}       
\usepackage{amsfonts}       
\usepackage{nicefrac}       
\usepackage{microtype}      
\usepackage{algorithmic}
\usepackage{algorithm}
\usepackage{amsmath}
\usepackage{graphicx}
\usepackage{caption}
\usepackage{amsmath}
\newtheorem{theorem}{Theorem}

\usepackage{pgfplots}
\pgfplotsset{compat=1.14}
\usepackage{wrapfig}

\usepackage{mathtools}

\usepackage[title]{appendix}
\usepackage{caption,subcaption}

  \title{GRASPEL: \underline{Gra}ph \underline{Spe}ctral  \underline{L}earning at Scale}

\author{Yongyu Wang \\
Michigan Technological University\\
\texttt{yongyuw@mtu.edu} \\
\And
Zhiqiang Zhao\\
Michigan Technological University\\
\texttt{qzzhao@mtu.edu} \\
\AND
Zhuo Feng \\
Stevens Institute of Technology\\
\texttt{zhuo.feng@stevens.edu} \\
}

\begin{document}

\maketitle

\begin{abstract} 
  Graph learning plays important roles in many data mining and machine learning tasks, such as manifold learning, data representation and analysis, dimensionality reduction, data clustering, and visualization, etc. In this work, for the first time we present a highly-scalable \emph{spectral graph densification} approach (GRASPEL) for  graph learning from data. By limiting the precision matrix to be a graph-Laplacian-like matrix in graphical Lasso, {our approach aims to learn ultra-sparse  undirected graphs from potentially  high-dimensional input data. A very unique property of the graphs learned by GRASPEL is that the spectral  embedding (or approximate effective-resistance)  distances on the graph will encode  the similarities between the original input data points. By  interleaving the latest high-performance nearly-linear time  spectral methods for   graph sparsification,  coarsening and embedding, ultra-sparse yet spectrally-robust graphs can be learned by identifying and including the most spectrally-critical edges into the graph.} Compared with prior state-of-the-art graph learning approaches, GRASPEL is more scalable and   allows   substantially improving computing efficiency and solution quality of a variety of data mining and machine learning applications, such as manifold learning, spectral clustering (SC), graph recovery, and dimensionality reduction. {For example,  when comparing with graphs constructed using prior methods,  GRASPEL achieved the state-of-the-art performance for spectral clustering and graph recovery. }
\end{abstract}

\section{Introduction}\label{sect:introduction}
Graph construction is playing increasingly important roles in many machine learning and data mining applications. For example, a key step of many existing machine learning methods requires converting potentially high-dimensional data sets into graph  representations: it is a common practice to represent each (high-dimensional) data point  as a  node,  and assign each edge a weight to encode the similarity between the two nodes (data points). The constructed graphs can be efficiently leveraged to represent  the underlying structure of a data set or the relationship between data points     \citep{jebara2009graph,maier2009influence,liu2018learning}. However, how to  learn meaningful graphs from large data set at scale still remains a  challenging problem.

Several recent graph learning methods  leverage emerging graph signal processing (GSP) techniques for estimating sparse graph Laplacians, which show very promising results     \citep{dong2016learning,egilmez2017graph,dong2019learning, kalofolias2017large}. For example,     \citep{egilmez2017graph}  addresses the graph learning problem by restricting the precision matrix  to be a graph Laplacian and maximizing a posterior estimation of Gaussian Markov Random Field (GMRF), while an $l1$-regularization term is used to promote graph sparsity;     \citep{rabbat2017inferring} provides an error analysis for inferring sparse graphs from smooth signals; { \citep{kalofolias2017large} leverages approximate nearest-neighbor (ANN) graphs to reduce the number of variables for optimization}. However,  even the state-of-the-art Laplacian estimation methods for graph learning do not scale well for large data set due to their extremely high algorithm complexity. For example, solving the optimization problem for Laplacian estimation  in  \citep{dong2016learning,kalofolias2016learn, egilmez2017graph,dong2019learning} requires  $O(N^2)$ time complexity per iteration for $N$ data entities and  nontrivial parameters tuning for controlling graph sparsity which limits their applications to only very small data sets (e. g. with up to a few thousands of data points). {The latest graph learning approach   \citep{kalofolias2017large} takes advantages of ANN graphs but can still run rather slowly for large data sets.}

{This work for the first time   introduces a \emph{spectral graph densification}  approach (GRASPEL) for  learning ultra-sparse graphs from data by leveraging the latest results in spectral graph theory \citep{feng2016spectral,feng2018similarity,zhao2018nearly}. There is  a  clear connection between our approach and the GSP-based Laplacian estimation methods \citep{dong2016learning,kalofolias2016learn, egilmez2017graph,kalofolias2017large, dong2019learning}. GRASPEL  also has a clear connection with the original graphical Lasso method \citep{friedman2008sparse}  with the precision matrix replaced by a  Laplacian-like matrix. Specifically, by treating $p$-dimensional data points as $p$ graph signals, GRASPEL learns a graph Laplacian   by maximizing the first few Laplacian eigenvalues as well as the smoothness of graph signals across edges, subject  to a  spectral stability constraint. }  A very unique property of the graphs learned by GRASPEL is that the spectral   embedding (or approximate effective-resistance) distances on the graph will encode  the similarities between the original input data points. 

{GRASPEL  can efficiently identify and include the most spectrally-critical edges into the latest graph by leveraging recent nearly-linear time spectral  sparsification, coarsening and embedding methods \citep{feng2016spectral,feng2018similarity,zhao2018nearly}. The iterative graph learning procedure will be terminated when the graph spectra become sufficiently stable (or graph signals become sufficiently smooth across the graph and lead to rather small Laplacian quadratic forms). Comparing with state-of-the-art methods, GRASPEL allows more scalable estimation of attractive Gaussian Markov Random Fields (GMRFs)  for even very large data set. We show through extensive experiments that GRASPEL can learn  high-quality ultra-sparse   graphs that  can be immediately leveraged to   significantly improve the    efficiency and accuracy of spectral clustering (SC) and graph recovery (GR) tasks;   the proposed approach also allows the development  of a multilevel t-Distributed Stochastic Neighbor Embedding (t-SNE) algorithm, showing a substantial runtime  improvement over existing methods    \citep{maaten2008visualizing,van2014accelerating}.  }


\section{Background of Graph Learning via Laplacian Estimation} 

Given $M$ observations on $N$ data entities stored in a data matrix $X\in {\mathbb{R} ^{N\times M}}$, each column of $X$ can be considered as a signal on a graph. The recent  graph learning method \citep{dong2016learning} aims to estimate a graph Laplacian from $X$ while achieving the following desired characteristics:

 \textbf{Smoothness of  Graph Signals. } {The graph signals corresponding to the real-world data should  be sufficiently smooth on the learned graph structure: the signal values will only change gradually across connected neighboring nodes.
The smoothness of a   signal $x$ over a   undirected graph   $G=(V,E, w)$ can be measured {with} Laplacian quadratic form 
\begin{equation}
{x^T}L x= \sum\limits_{\left( {p,q} \right) \in E}
{{w_{p,q}}{{\left( {x\left( p \right) - x\left( q \right)}
\right)}^2}},
\end{equation}

where $L=D-W$ denotes the Laplacian  matrix of  graph  $G$ with $D$ and $W$ denoting the  degree  and   the weighted adjacency matrices of  $G$, and $w_{p,q}=W(p,q)$  denotes  the weight for edge ($p,q$).  The smaller value of quadratic form indicates the smoother   signals across the graph. It is also possible to quantify  the smoothness ($Q$) of a set of signals $X$ over graph $G$  using the following matrix trace   \citep{kalofolias2016learn}:
\begin{equation}
Q(X,L)=Tr({X^T L X}),
\end{equation}
where $Tr$ denotes the matrix trace. }

 \textbf{Sparsity of the Estimated Graph (Laplacian).} {Graph sparsity   is another critical consideration in graph learning. One of the most important motivations of learning a graph is to use it for downstream data mining or machine learning tasks. Therefore,   desired graph  learning algorithms should allow better capturing and  understanding the  global structure (manifold) of the data set, while producing sufficiently sparse graphs  that   can be easily stored and efficiently manipulated in the downstream  algorithms, such as graph clustering, partitioning, dimension reduction, data visualization, etc. To this end,  the graphical Lasso algorithm \citep{friedman2008sparse}  has been proposed to learn the structure in an undirected Gaussian graphical model  using $l1$ regularization to
control the sparsity of the precision matrix. Given a sample covariance matrix $S$ and a regularization parameter $\beta$, graphical Lasso targets the following convex optimization task:
\begin{equation}\label{formula_lasso}
\max_{\Theta}: \log\det(\Theta)-  Tr(\Theta S)-{\beta}{{\|\Theta\|}}^{}_{1} ,
\end{equation}
over all non-negative definite   precision matrices $\Theta$. The first two terms together can be interpreted as the log-likelihood under a Gaussian Markov Random Field (GMRF). $\|\cdot\|$ denotes the entry-wise $l1$ norm, so ${\beta}{{\|\Theta\|}}^{}_{1} $ becomes the sparsity promoting regularization term. This model tries to learn the graph structure by maximizing the penalized log-likelihood. If the covariance matrix $S$ is obtained by sampling a $N$-dimensional Gaussian distribution
with zero mean, each element in the precision matrix $\Theta_{i,j}$ encodes the conditional dependence between variables $X_i$ and $X_j$. For example, $\Theta_{i,j}=0$ implies that the corresponding variables $X_i$ and $X_j$
are conditionally independent, given the rest. However, the log-determinant problems are very computationally expensive. The emerging GSP-based methods infer the graph by adopting the criterion of signal smoothness \citep{kalofolias2016learn,dong2016learning,egilmez2017graph, kalofolias2017large}. However, their extremely high complexities  do  not allow for learning  large-scale graphs  involving {millions or even hundred thousands  of nodes}. Furthermore, these methods usually require nontrivial parameters tuning for controlling graph sparsity.}

\section{GRASPEL: Graph Spectral Learning at Scale}

At high level, GRASPEL  gains insight from recent GSP-based Laplacian estimation methods \citep{dong2019learning}, aiming to solve the following  convex optimization problem that is similar to the graphical Lasso problem \citep{friedman2008sparse}:
\begin{equation}\label{opt2}
  {\max_{\Theta}}: \log\det(\Theta)-  \frac{1}{M}Tr({X^T \Theta X})-\beta {{\|\Theta\|}}^{}_{1},
\end{equation}
where   $\Theta={L}+\frac{I}{\sigma^2}$, ${L}$ denotes the set of valid graph Laplacian matrices, $I$ denotes the identity matrix, and $\sigma^2>0$ denotes prior feature variance. It can be shown that the three terms in (\ref{opt2}) are corresponding to  $\log\det(\Theta)$, $Tr(\Theta S)$ and  ${\beta}{{\|\Theta\|}}^{}_{1}$ in (\ref{formula_lasso}),  respectively. When  each   column vector  in the  data matrix $X$ 
\footnote{For each of the original $M$-dimensional  data vectors $X(i,:)\in {\mathbb{R} ^{1\times M}}$ where $i=1,...,N$, the following pre-processing step will be performed: $X(i,:)= {X(i,:)-\mu_i}$ where $\mu_i$ denotes the sample mean of $X(i,:)$; as the result, the sample  covariance matrix $S$ in (\ref{formula_lasso}) becomes $\frac{XX^{\top}}{{ {M}}}$. } 
is treated as a graph signal vector, there is a close connection between our formulation and the graphical Lasso problem. Since $\Theta={L}+\frac{I}{\sigma^2}$ matrices correspond to  symmetric and positive definite  (PSD) matrices (or M matrices) with non-positive off-diagonal entries, this formulation will lead to the estimation of  attractive GMRFs \citep{dong2019learning}.


\subsection{ Theoretical Background}\label{sec:theory}
 Express the Laplacian matrix as
\begin{equation}\label{LaplacianEdge}
L=\sum\limits_{\left( {p,q} \right) \in E}
{w_{p,q}}e_{p,q}e^\top_{p,q}
\end{equation}
where  ${e_{p}} \in \mathbb{R}^N$ denotes  the standard basis vector with all zero entries except for the $p$-th entry being $1$, and    ${e_{p,q}}=e_p-e_q$. Considering the cost function $F$  in (\ref{opt2}):
\begin{equation}\label{optF0}
F= \log\det(\Theta)-  \frac{1}{M}Tr({X^T \Theta X})-\beta {{\|\Theta\|}}^{}_{1},
\end{equation}
the partial derivative with respect to the weight $w_{p,q}$ of edge $(p,q)$ can be written as:
\begin{equation}\label{optF}
\frac{\partial F}{\partial w_{p,q} } = \sum\limits_{i=2}^N \frac{1}{\lambda_i+1/\sigma^2} \frac{\partial \lambda_i}{\partial w_{p,q} }
-  \frac{1}{M}\|X^\top e_{p,q}\|_2^2-\beta,
\end{equation}
where  the Laplacian eigenvectors corresponding to the ascending eigenvalues  ${\lambda _i}$ are denoted by ${u_i}$ for $i=1,...,N$, satisfying:
\begin{equation}\label{eigen}
L u_i = \lambda_i u_i.
\end{equation}
According to the first-order spectral perturbation analysis   in Theorem \ref{thm:pertub} (Appendix), we have:
\begin{equation}\label{edgePerturb}
\frac{\partial \lambda _i}{\partial w_{p,q} }=\left( {{{u_i^T}e_{p,q}} } \right)^2.
\end{equation}
Construct a  subspace matrix for spectral graph embedding using the first $r-1$ weighted nontrivial  Laplacian eigenvectors as follows:
\begin{equation}\label{subspace}
U=\left[\frac{u_2}{\sqrt {\lambda_2 +1/\sigma^2}},..., \frac{u_r}{\sqrt {\lambda_r +1/\sigma^2}}\right]. 
\end{equation}
Then (\ref{optF}) can be approximately written as follows 
\begin{equation}\label{optF2}
\frac{\partial F}{\partial w_{p,q} } \approx \|U^\top e_{p,q}\|_2^2
-  \frac{1}{M}\|X^\top e_{p,q}\|_2^2-\beta.
\end{equation}
\subsection{Graph Learning via Spectral   Densification}
 \textbf{Spectrally-critical edges.} Define    {spectrally-critical edges} to be the ones that can most effectively perturb the graph spectral properties, such as the first few Laplacian eigenvalues and eigenvectors. Then $\|U^\top e_{p,q}\|_2^2$  can be interpreted  as the spectral sensitivity of the  candidate edge ($p, q$): adding a candidate edge ($p, q$) with larger $\|U^\top e_{p,q}\|_2$  into the latest graph will more significantly perturb the first few Laplacian eigenvalues and eigenvectors.   (\ref{optF2})  implies that  spectrally-critical edges will also significantly impact the objective function (\ref{optF0}) and  have large \textit{spectral embedding distortions}  that are defined as 
\begin{equation}\label{embedDist}
\eta_{p,q}=\frac{z_{p,q}^{emb}}{z_{p,q}^{data}},
\end{equation}
where $z_{p,q}^{emb}=\|U^\top e_{p,q}\|^2_2$ and $z_{p,q}^{data}=\frac{\|X^\top e_{p,q}\|^2_2}{M}$ denote the $l2$ distances in the spectral embedding space and the  original data vector space (averaged among $M$ samples), respectively. Not surprisingly, as $\sigma^2$ and $r$ in (\ref{subspace}) approach $+\infty$, $z_{p,q}^{emb}$ becomes the effective-resistance distance, and $\eta_{p,q}$  becomes the edge leverage score for spectral graph sparsification \citep{spielman2011graph} when each edge weight is computed by ${w_{p,q}}=\frac{1}{z_{p,q}^{data}}$. 
Subsequently, the partial derivative in (\ref{optF2}) can be further simplified as follows:
\begin{equation}\label{optF3}
 \frac{\partial F}{\partial w_{p,q} } \approx \left(1-\frac{1}{\eta_{p,q}} \right) z_{p,q}^{emb}-\beta,
\end{equation}
implying that  as long as $\eta_{p,q}>1$ holds for the edge $(p,q)$, the greater $\eta_{p,q}$  and $z_{p,q}^{emb}$ values will result in more significant improvement (increase) of the objective function.

 \textbf{Spectral graph densification.} Prior research proves that every undirected graph has a nearly-linear-sized spectral sparsifier that can be computed in nearly-linear time by sampling each edge with a probability proportional to its  leverage score \citep{spielman2011graph}; on the contrary, the proposed  GRASPEL framework can be considered as a {spectral graph densification} procedure that aims to identify and include the edges with large (effective-resistance) embedding distortions. Therefore, the goal of graphical Lasso under   Laplacian-like precision matrix constraint is equivalent to identifying  spectrally-critical edges that can most effectively decrease the distortion in the spectral embedding space. As a result, a very important feature of GRASPEL  is that the effective resistances     on the learned graph   will approximately encode the Euclidean ($l2$) distances between the original data points. 

   
\textbf{Convergence analysis.}   The global maximum can be obtained when (\ref{optF2}) becomes zero. The convergence  can be controlled by properly setting an embedding distortion threshold for $\eta $ or sparsity constraint parameter $\beta$. When $\beta=0$, the global maximum can be obtained when $\|U^\top e_{p,q}\|_2^2= \frac{1}{M}\|X^\top e_{p,q}\|_2^2$, which means the spectral (effective-resistance) embedding distances on the graph exactly match the corresponding Euclidean distances between the original data points. However, it is well-known that  choosing a proper $\beta$ value for graphical Lasso can be very challenging in practical applications. 
      On the other hand, the convergence of GRASPEL can  be determined based on \emph{graph spectral stability}: when the maximum effective-resistance or spectral embedding distortion $\eta$ becomes  small enough, or the first few Laplacian eigenvalues and eigenvectors become sufficiently stable, the GRASPEL optimization iterations can be terminated. Compared with the prior graphical Lasso method that  always faces troubles  finding the most suitable sparsity constraint parameter $\beta$, GRASPEL requires a much simpler converge control scheme.

\subsection{Overview of the GRASPEL Framework}\label{sec:overview}
\begin{figure*}\centering
\includegraphics[width=0.9975\textwidth]{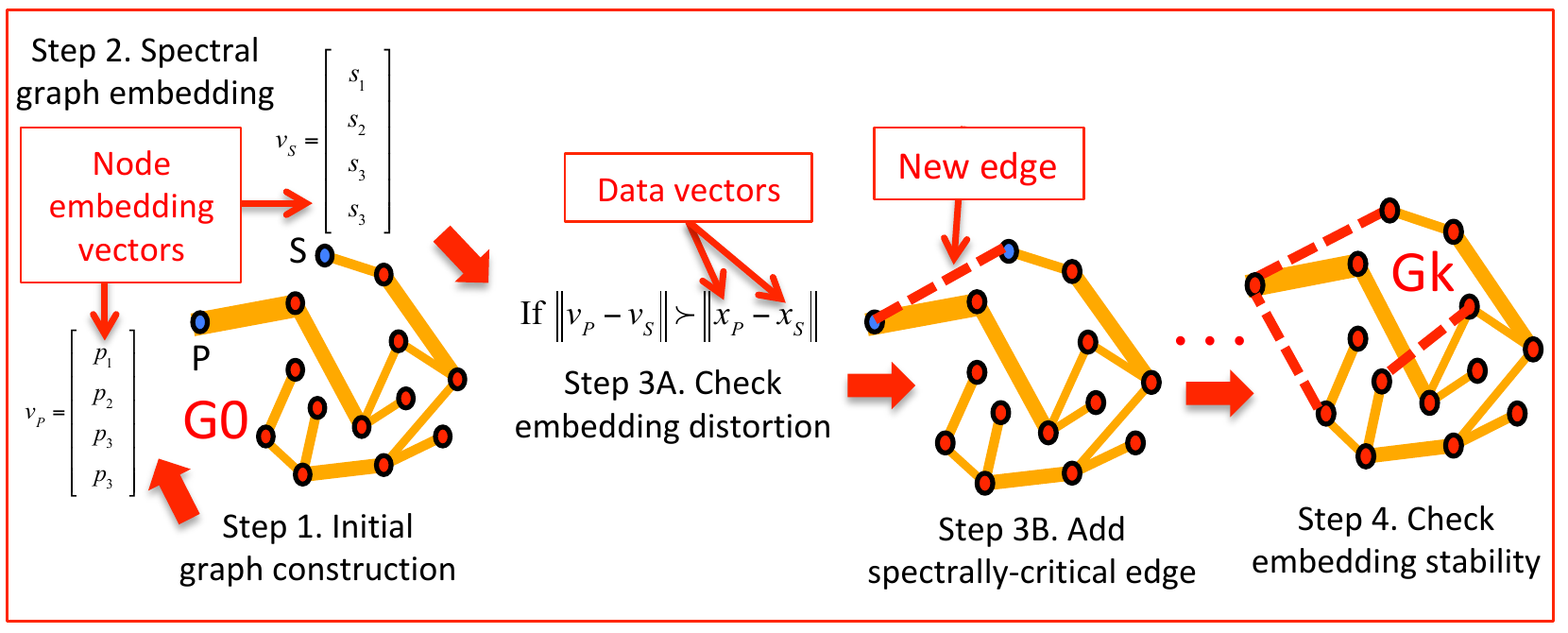}
\caption{The overview of the proposed GRASPEL framework.\protect\label{fig:overview}}
\end{figure*}
 \textbf{Key steps in GRASPEL.} To achieve good efficiency in graph learning that may involve large number of nodes,  GRASPEL leverages a  spectral approach for   solving  (\ref{opt2}) efficiently. GRASPEL aims  to iteratively identify and add the most spectrally-critical edges  into the latest graph until  no such edges can be found, which consists of the following key steps as illustrated in Figure \ref{fig:overview}: 
\begin{itemize}

\item \textbf{Step (1): Initial graph construction.} {Similar to \citep{kalofolias2017large}, we start  with constructing an ANN graph, and subsequently convert it into an ultra-sparse nearest-neighbor (uNN) graph with $O(N\log N)$   edges by  leveraging a nearly-linear-time spectral sparsification algorithm \citep{feng2018similarity}.}

\item 
 \textbf{Step (2): Spectral graph embedding.} {We apply a  nearly-linear-time spectral  embedding procedure \citep{zhao2018nearly}} to the current graph   so that each node  will be associated with a low-dimensional embedding vector (e.g. $v_s$ for node $S$ in Figure \ref{fig:overview}).

\item 
\textbf{Step (3): Spectrally-critical edge identification.} We  quickly identify the edges with the largest {embedding distortion} and include them into the latest graph.  

\item 
 \textbf{Step (4): Spectral stability checking. } After repeating the  {Steps (2)-(3)} multiple times for adding new edges, GRASPEL  will return the final  graph  {if the spectral embedding distortion is sufficiently small.} 
\end{itemize}

 \textbf{Complexity analysis.} {To achieve scalable spectral graph embedding key to identification of spectrally-critical edges in Steps (2)-(3), we will leverage the latest high-performance spectral   sparsification \citep{spielman2011graph,feng2019grass},     spectral coarsening \citep{loukas2018spectrally,loukas2019graph}, and spectral     embedding \citep{zhao2018nearly} algorithms. Since all the kernel functions involved in GRASPEL, such as ANN graph construction \citep{muja2009fast,muja2014scalable,malkov2018efficient}, are implemented based on nearly-linear $O(N \log N)$ time algorithms, the entire spectral graph learning approach  GRASPEL also has a nearly-linear time complexity.  }
 


\section{Detailed Steps in GRASPEL}

{\textbf{Initial graph construction.}}
 As aforementioned, (approximate) kNN graphs can be used to construct the initial graphs in Step (1),  since they can be created very efficiently \citep{muja2009fast}, while being able to approximate the local data proximity \citep{roweis2000nonlinear}. However, traditional kNN graphs have the following drawbacks: \textbf{1)} The kNN graphs with large $k$ (the number of nearest neighbors) has the tendency of increasing the cut-ratio  \citep{qian2012graph}; \textbf{2)} The optimal $k$ value is usually problem dependent and can be very difficult to find. In this work, we will start creating an (approximate) kNN graph with a relatively small $k$ value (e.g. $k=2$ ), and strive to significantly improve the graph quality by adding extra spectrally-critical edges through implicitly solving the proposed optimization problem in (\ref{opt2}). In addition, uNN graphs can be exploited via spectral sparsification   \citep{ spielman2011graph,feng2019grass}   to further simplify the kNN graph \citep{wang2017towards}.

\textbf{Spectral graph embedding.}
Spectral graph embedding directly leverages the first few nontrivial eigenvectors  for mapping nodes onto low-dimensional space \citep{belkin2003laplacian}. The eigenvalue decomposition of  Laplacian matrix  is usually the computational bottleneck in spectral graph embedding, especially for large graphs \citep{shi2000normalized, von2007tutorial,chen2011parallel}. To achieve good scalability for computing the first few Laplacian eigenvalues,  we can exploit fast Laplacian solvers \citep{miller:2010focs} or multilevel Laplacian solvers that allow for much faster eigenvector (eigenvalue) computations   without loss of accuracy \citep{zhao2018nearly}. 


\textbf{Spectrally-critical edge identification.}
Once Laplacian eigenvectors are available for the current graph,
we can identify spectrally-critical edges by looking at each candidate edge's  embedding distortion defined in (\ref{embedDist}).
To this end, we exploit the following first-order spectral
perturbation analysis to quantitatively evaluate each candidate edge's impact on the first few eigenvalues.   The following theorem will allow us to identify the most spectrally-critical edges leveraging the first few Laplacian eigenvectors.
  
\begin{theorem}\label{thm:pertub}
The spectral criticality $c_{p,q}$ or embedding distortion $\eta_{p,q}$ of a candidate edge   $({p,q})$  on the Laplacian eigenvalue ${\lambda_i}$ can be properly estimated by $c_{p,q}= w_{p,q}\left( {{{u_i^T}e_{p,q}}  } \right)^2 \propto  {\eta_{p,q}}= \frac{z^{emb}_{p,q}}{z^{data}_{p,q}}$.
\end{theorem}
Proof: See the Appendix.

\emph{{Edge identification with multiple eigenvectors.}} With $k$ eigenvectors for spectral embedding, we can first project the graph nodes onto a $k$-dimensional space and perform spectral clustering to group  the nodes into $k$ clusters, {where the  embedding dimension $k$  can be determined based on the largest gaps of the first few (e.g. $100$) Laplacian eigenvalues \citep{peng2015partitioning}.} Next, according to (\ref{optF3}) we only have to examine the candidate   edges that connect nodes between two distant clusters in the embedding space,  and sort them based on  embedding distortions.  However,  it can be very difficult in practice to choose a proper embedding dimension $k$ for general graph learning tasks.  

\emph{Edge identification with Fiedler vectors.}  Our approach for identifying spectrally-critical edges  starts with sorting nodes according to the Fiedler vector that can be computed in nearly-linear time leveraging fast Laplacian solvers \citep{miller:2010focs,spielman2014nearly}. This scheme is equivalent to including only the first nontrivial Laplacian eigenvector into the subspace matrix in (\ref{subspace}) for spectral   embedding, which allows     approximating the  gradient in the proposed optimization task (\ref{optF}). Since the distances computed using the Fiedler vector will be the lower bounds of the effective-resistance distances, the lower bound of  embedding distortions due to such an approximation can be efficiently estimated. According to (\ref{optF3}), only a small portion of node pairs with   large  embedding distances needs to be examined as candidate edges. Consequently, we can limit the search within the candidate edge  connections between the top and bottom few nodes in the  1D sorted node vector. Only the candidate edges with  top   embedding distortions will be added into the latest graph. The algorithm  for spectrally-critical edge identification using the Fiedler vector  has been  described in  Algorithm \ref{alg:graspel} in the Appendix. 

\emph{{Multilevel edge identification with spectral  coarsening.}} To further reduce the computational cost for large data sets, it is possible to perform spectral graph  coarsening \citep{zhao2018nearly,loukas2018spectrally,loukas2019graph} and subsequently search for high-distortion candidate edges on the coarsest graph. Once  a small set of top spectrally-critical edges has been identified, we will find their corresponding candidate edges in the original graph. Since each coarse-level candidate edge may correspond to multiple  candidate edges in the original graph, we will sort them based on their embedding distortions, and only add the ones with largest distortions into the latest graph.  




\begin{wrapfigure}{R}{2.85025002in}
 \vspace*{-13pt}
 \hspace*{-5pt}\includegraphics[width=2.85in]{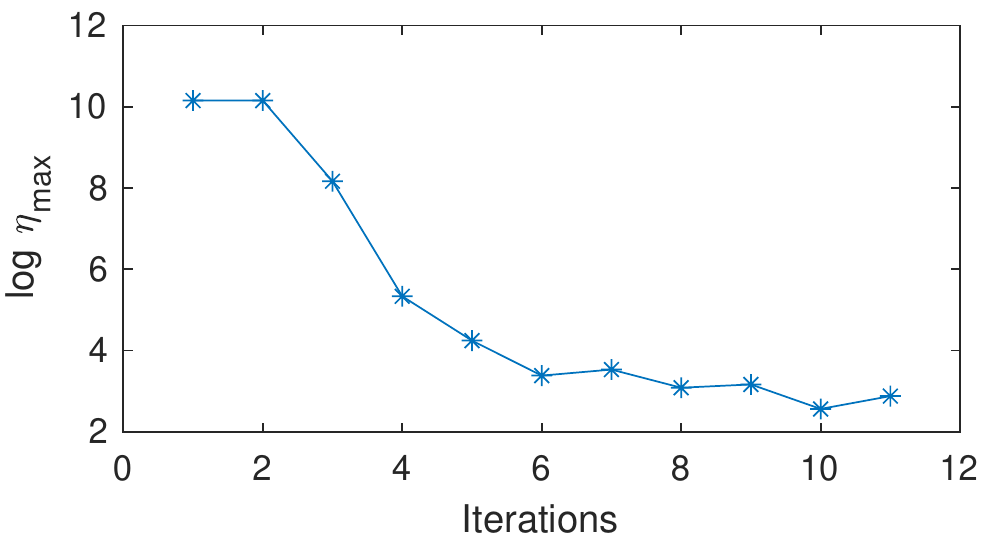}
 \vspace*{-5pt}
 \caption{\small The decreasing embedding distortions.\protect\label{fig:convergance}}
 \vspace*{-5pt}
 \end{wrapfigure}
\textbf{Spectral stability checking.} The following naive scheme can be exploited for checking the spectral stability of each graph learning iteration: \textbf{1)} In each iteration, we compute and record the several smallest eigenvalues of the latest graph Laplacian according to the largest gap between eigenvalues \citep{peng2015partitioning}: for example, the first (smallest) $k$ nonzero eigenvalues that are critical for spectral clustering   will be stored; \textbf{2)} We check whether   sufficiently stable  spectra have been reached for graph learning by comparing them with the eigenvalues computed in the previous iteration: if the change   is significant, more iterations may be needed. 

However, finding a proper spectral gap in the first few eigenvalues can be quite tricky in practice, not to mention the potentially high computational cost. In this work, we propose to evaluate the  edge embedding distortions with (\ref{embedDist}) for checking the spectral stability of the learned graph. If there exists no additional edge that has an embedding distortion greater than a given tolerance level, GRASPEL iterations can be terminated. It should be noted that choosing different  tolerance levels  will result in  graphs with  different densities. For example, choosing a smaller distortion tolerance will require more edges   to be included so that the resultant spectral embedding distances on the learned graph can more precisely encode the distances between the original data points. In practice, GRASPEL converges very quickly even for very large data set. As shown in Figure \ref{fig:convergance},  starting from an initial 2NN graph of the USPS data set, GRASPEL adds $0.001 |V|$ additional edges in each iteration and requires $11$ iterations to effectively mitigate the maximum embedding distortion by over $2,600\times$.


\section{Experiments}
In this section, extensive experiments have been conducted to evaluate the performance of GRASPEL (using Fiedler vectors) for a variety of public domain data sets (see the Appendix \ref{dataApp} for detailed setting and evaluation metrics). In Sections \ref{sec:GR} and \ref{sec:DR} (see Appendix), we report additional experimental results for graph recovery and dimensionality reduction applications leveraging the proposed GRASPEL approach. When checking the spectral distortion of candidate edges, we randomly sample among the data points that correspond to the top and bottom $0.05|V|$ ($\epsilon=0.05$ in Algorithm \ref{alg:graspel}) of the sorted node array according to the Fiedler vector, which allows quickly identify the most spectrally-critical edges. Each GRASPEL iteration will add $0.001|V|$ ($\zeta=0.001$ in Algorithm \ref{alg:graspel}) additional edges into the latest graph. $\sigma = 1E3$ in (\ref{subspace}) is used for all experiments. {Note that all the graphs learned by GRASPEL have ultra-sparse  structures with relatively low graph densities (defined as $|E|/|V|$); GRASPEL allows learning much sparser graphs when comparing with the latest approach that produces much denser graphs \citep{kalofolias2017large}. }
\subsection{Spectral Stability Checking}

\begin{figure*}\centering
\includegraphics[width=0.995\textwidth]{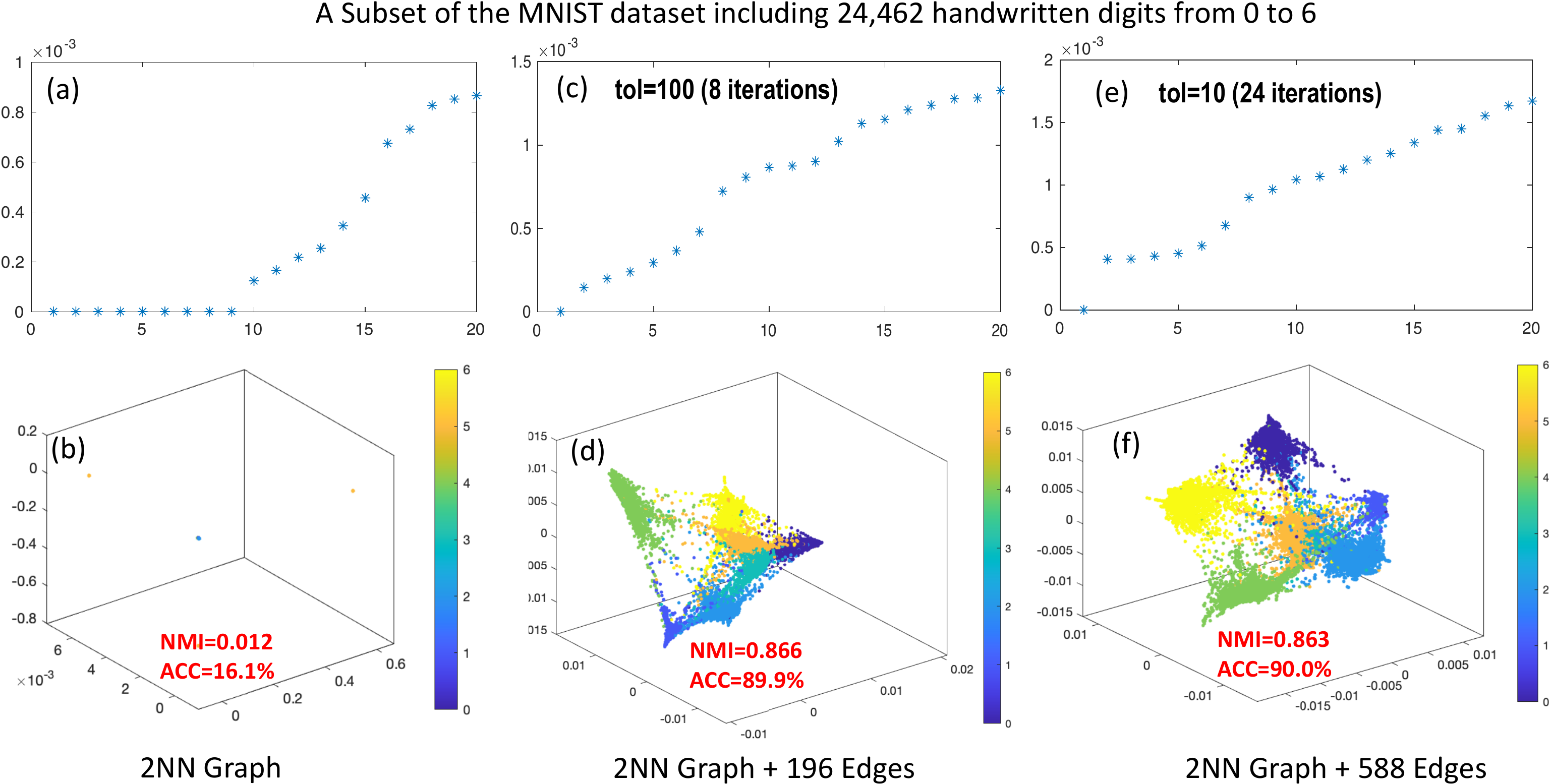}
\caption{The first $20$ Laplacian eigenvalues (top) and  spectral  drawings (bottom) of the 2NN graph in figures (a) and (b), and the GRASPEL-learned graphs  in figures (c) to (f). \protect\label{fig:EigResultMNIST}}
\end{figure*}
We first show how the proposed scheme for spectral stability checking can be applied based on the  embedding distortion metric defined in (\ref{embedDist}). The proposed graph learning iterations will be terminated when there exists no candidate edge that has a spectral embedding distortion greater than a given tolerance level (e.g. $\eta\ge tol$).  In Figure \ref{fig:EigResultMNIST}, we show the   first $20$ Laplacian eigenvalues (top figures) and  spectral  drawings (bottom figures) of the graphs learned with different  distortion tolerance levels for a subset ($24,462$ handwritten digits from $0$ to $6$) of the MNIST data set (see Appendix for details). When producing the spectral drawing layouts, each entry of the first three nontrivial Laplacian eigenvectors ($u_2, u_3, u_4$) is used as   the $x-$, $y-$ and   $z-$coordinate  of each node (data point), respectively. The ground-truth label of each data point is also shown in different colors. The edges are not shown in the layouts for more clearly illustrating the clusters of the data points.

\begin{wrapfigure}{R}{2.85025002in}
 \vspace*{-3pt}
 \hspace*{-5pt}\includegraphics[width=2.85in]{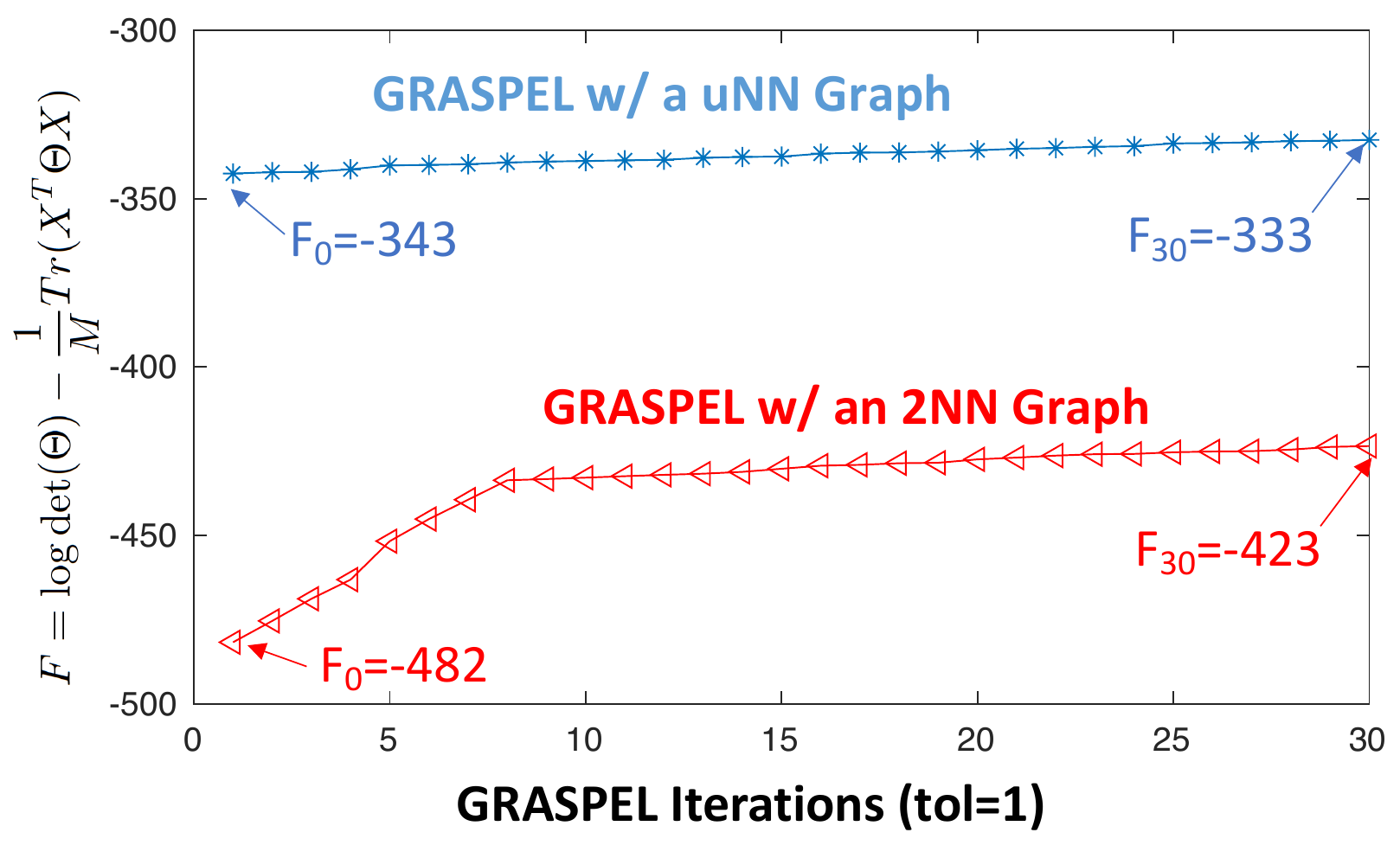}
 \vspace*{-5pt}
 \caption{\small  GRASPEL with 2NN and uNN graphs. \protect\label{fig:objectiveFunction}}
 \vspace*{-5pt}
 \end{wrapfigure}
Two embedding distortion tolerance levels ($tol=100$ and $tol=10$) are considered in our experiment. Starting from an initial 2NN graph ($\frac{|E|}{|V|}=1.567$), by adding $0.001 |V|$ edges in each iteration the $tol=100$   option requires  $8$ iterations ($11$ seconds), whereas the $tol=10$ option requires $24$ iterations ($36$ seconds) to converge.  As observed, the lower tolerance option produces  a slightly denser graph and also a more closely coupled spectral embedding result when comparing with the results produced with $tol=100$. Not surprisingly,  GRASPEL    has dramatically mitigated the spectral embedding distortions from $\eta_{max}=1E6$ to $\eta_{max}=10$ by adding only $0.024 |V|$ ($588$) additional edges into the initial 2NN graph. By examining the first few Laplacian eigenvalues, we notice that the initial 2NN has nine connected components (that equals the number of zero eigenvalues), while with $196$ extra edges added via GRASPEL iterations, a well-connected graph can be formed for approximately preserving the structure of the original data set. The gaps between the $7$th and $8$th eigenvalues in (c) and (d)  indicate  that the dimensionality for both GRASPEL-learned graphs   is approximately six.  

\subsection{Convergence of GRASPEL Iterations}
We also show how the objective function defined in (\ref{optF0}) would change during the GRASPEL iterations when starting with \textbf{(a)} a 2NN graph, and  \textbf{(b)} a uNN graph for the USPS data set. Only the first $50$ Laplacian eigenvalues are computed for evaluating (\ref{optF0}). The uNN graph is obtained by  spectrally  sparsifying a 5NN graph using the GRASS algorithm (with a relative condition number of $30$) \citep{feng2019grass}. We set the distortion tolerance $tol=1$ for both cases and demonstrate the results for the first thirty GRASPEL iterations. As observed in Figure \ref{fig:objectiveFunction}, comparing with (b), (a) achieves a much greater objective function value  after $30$ iterations, not to mention the lower density in the learned graph: ($1.33$  vs $1.59$). Therefore, for very large data sets before going through the GRASPEL iterations   a spectral sparsification procedure would be indispensable for   achieving  substantially improved solution quality and runtime efficiency. When starting with the 2NN graph,      (\ref{optF0}) grows slowly after $10$ iterations, indicating  rather small gradient values since there exist too few edges with large embedding distortions, which   justifies an earlier termination of the GRASPEL iterations.
 
\subsection{Graph Learning for  Spectral Clustering (SC)}
The  classical spectral clustering (SC) algorithm (see Algorithm \ref{alg:sc} in the Appendix) first constructs a graph  where each edge weight encodes similarities between different data points (entities); then SC calculates the eigenvectors of the graph Laplacian matrix and embeds data points into low-dimensional space \citep{belkin2003laplacian}; in the last, k-means algorithms are used to partition the data points into multiple clusters. The performance of SC strongly depends on the quality of the underlying graph     \citep{guo2015robust}.   In this section, we apply GRASPEL for graph construction, and show the learned graphs can result in drastically improved efficiency and accuracy in SC tasks. 


\begin{table*}[htb]
\begin{center}
\scriptsize
\addtolength{\tabcolsep}{-2.5pt}\centering
\caption{{Spectral Clustering Results}}
\begin{tabular}{ |c|c|c|c|c|c|c|c|c|c|c|c|c|c|c|c|}
\hline
 &\multicolumn{4}{|c|}{ACC(\%)/ NMI/ Time-C (seconds)/Time-S (seconds)/Graph Density (GRASPEL)} \\
 \hline Data Set&Standard KNN &ConskNN &{ LSGL }&GRASPEL (Algorithm \ref{alg:graspel})\\
 \hline  COIL-20 &78.80/ 0.86/ \textbf{0.36}/0.37 & 79.86/ 0.86/ 0.54/0.28&{ 65.48/0.77/60.22/1.02}&\textbf{90.27}/ \textbf{0.96}/ 0.40/\textbf{0.19}/\textbf{1.19} \\
 \hline  PenDigits&81.12/ 0.80/\textbf{ 1.25}/0.47   &84.17/ {0.81}/ 8.59/46.41&{ 75.01/0.72/1622/15.64}&\textbf{85.96}/ \textbf{0.82}/ 4.51/\textbf{0.27}/\textbf{1.10}  \\
 \hline USPS &68.22/ 0.77/ \textbf{2.66}/1.02  &78.94/ 0.82/ 19.82/74.57&{ 72.35/0.71/2598/29.37}&\textbf{92.59}/ \textbf{0.87}/ 5.19/\textbf{0.21}/\textbf{1.10} \\    
 \hline  MNIST &71.95/ 0.72/ 242.38/6785&-&-&\textbf{81.67}/ \textbf{0.75}/ \textbf{59.27}/\textbf{2.90}/\textbf{1.10}\\
 
 \hline
\end{tabular}\label{table:clustering results}

        \footnotesize
        \item[-] indicates that the method is not capable for handling data sets of this scale.
        
\end{center}
\end{table*}
\textbf{SC with uNN graphs.} Table~\ref{table:clustering results} shows the ACC and NMI results of SC with graphs constructed by different  methods with the best numbers highlighted, {where graph construction time (Time-C) and spectral clustering time (Time-S) that involves eigendecomposition and kmeans clustering have also been reported.} We also report the graph densities ($\frac{|E|}{|V|}$) for the graphs learned by   GRASPEL when starting with uNN graphs computed via spectral sparsification procedures \citep{feng2019grass,spielman2011graph}. Note that the   high computational  and memory cost of recent GSP-based graph learning methods, such as GL-SigRep \citep{dong2016learning}, GL-Logdet \citep{dong2016learning} and GLSC \citep{egilmez2017graph} do not allow for processing data sets with more than a few thousands   data  points, thus can not be used for real-world SC tasks. As observed,  GRASPEL can consistently lead to dramatic performance improvement in SC, beating all   competitors in  clustering accuracy (ACC) across all  data sets: GRASPEL   achieves more than $18\%$ accuracy gain on USPS and $13\%$   gain on COIL20 over the second-best methods; for the MNIST data set GRASPEL also achieves over $14\%$ accuracy gain over the SC with standard kNN graph and more than $6X$ speedup in graph construction time. {Note that the graphs learned by GRASPEL (starting with uNN graphs) are ultra sparse, thereby allowing much faster eigendecompositions in SC  when comparing with other methods \citep{wang2017towards}: the SC of the MNIST data set with  standard kNN takes over $6,000$ seconds, which will be dramatically improved to require less than three seconds (over $2,000X$ speedup) using the graph learned by our method (GRASPEL).} 

\begin{figure*}\centering
\includegraphics[width=0.995\textwidth]{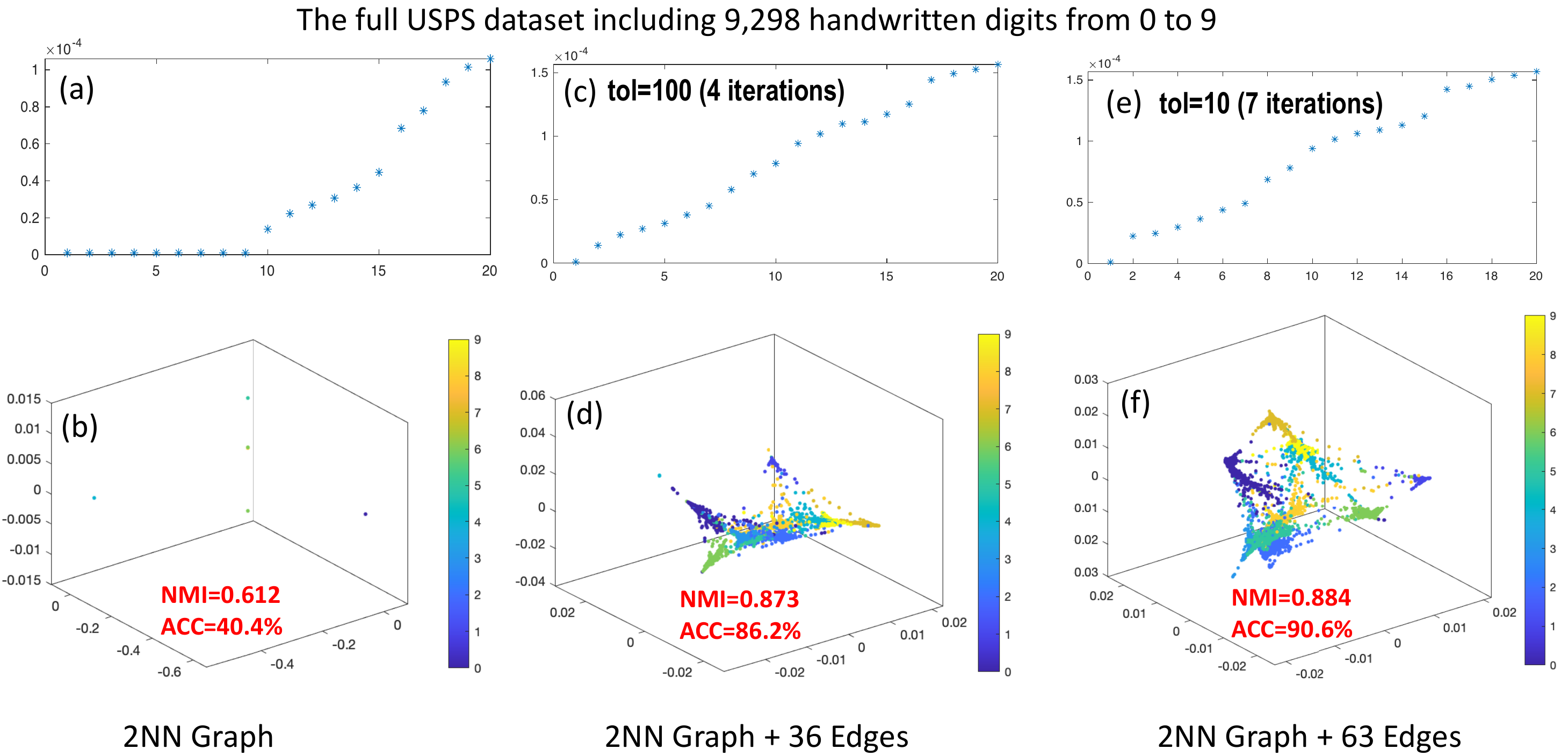}
\caption{The first $20$ Laplacian eigenvalues (top) and  spectral  drawings (bottom) of  the 2NN graph in figures (a) and (b), and the GRASPEL-learned graphs  in figures (c) to (f). \protect\label{fig:EigResultUSPS}}
\end{figure*}

\textbf{SC with 2NN graphs.} In Figure \ref{fig:EigResultMNIST} and Figure \ref{fig:EigResultUSPS}, additional SC results have been provided for the subset of the MNIST data set and the full USPS data set by comparing two embedding distortion tolerance levels ($tol=100$ and $tol=10$).    Not surprisingly, when starting with initial 2NN graphs a few GRASPEL iterations  have already dramatically improved SC results:  the normalized mutual information (NMI) was improved from $0.012$ to $0.863$ for the subset of the MNIST data set, and from $0.612$ to $0.884$ for the USPS data set; The clustering accuracy (ACC) was  improved from $16.1\%$ to $90.0\%$ for the subset of the MNIST data set, and from $40.4\%$ to $90.6\%$ for the USPS data set 

\textbf{Discussion.} The superior performance of GRASPEL is due to the following reasons: 
\noindent \textbf{1)} In traditional kNN graphs, all the nodes have the same degrees; as a result, the clustering  may strongly favor balanced cut, which may lead to improper cuts in high-density regions of the graph.  In contrast, GRASPEL always learns ultra-sparse graphs that only include  edges with the largest impact to graph spectral (structural) properties; as a result, the corresponding cuts   will always occur in proper regions of the graph, which enables to handle even unbalanced data. 
 \textbf{2)} Recent work \citep{garg2018spectral} shows the fundamental connections between spectral properties of graphs associated with data and the inherent robustness to adversarial examples. Since GRASPEL identifies candidate edges by leveraging spectral graph properties, the learned graph structure will also be robust to input noises (perturbations).

 \vspace{-10pt}
\section{Conclusion}
In this work,   we present a highly-scalable \emph{spectral graph densification} approach (GRASPEL) for  graph learning from data. By limiting the precision matrix to be a graph-Laplacian-like matrix in graphical Lasso, {our approach aims to learn ultra-sparse  undirected graphs from potentially  high-dimensional input data. By  interleaving the latest high-performance nearly-linear time  spectral methods for   graph sparsification,  coarsening and embedding, ultra-sparse yet spectrally-robust graphs can be learned by identifying and including the most spectrally-critical edges into the graph.} Compared with prior state-of-the-art graph learning approaches, GRASPEL is more scalable and   leads to   substantially improved computing efficiency and solution quality for a variety of data mining and machine learning applications, such as manifold learning, spectral clustering (SC), graph recovery, and dimensionality reduction.

 \section*{Broader Impact}
The success of the proposed approach for graph learning from data will significantly advance the state of the art in high-dimensional statistical data analysis, data mining, and machine learning,  leading to highly-scalable algorithms for manifold learning and dimensionality reduction,   graph recovery and data clustering, etc. The outcome of this research plan will be disseminated to  research communities via presentations and publications in  conferences and journals.  The  developed algorithms   will be disseminated for potential industrial adoptions. The resultant algorithms/software packages  will be available to  researchers and industrial partners.  The  proposed method  is also likely to influence general computer science and engineering fields related to complex system/network modeling, convex optimizations, numerical linear algebra, computational biology,  transportation and social networks, etc.
\bibliography{iclr2020,fengnew}
\bibliographystyle{iclr2020_conference}
\newpage
\appendix

\begin{appendices}

\section{Proof of Theorem 1 }

Let $L_{P}$  denote the Laplacian matrix of an undirected graph $P$, and $u_i$ denote the $i$-th eigenvector of $L_{P}$ corresponding to the $i$-th eigenvalue ${\lambda_i}$ that satisfies:
\begin{equation}\label{formula_eig_perturb0}
L_pu_i=\lambda_i u_i,
\end{equation}
then we have the following eigenvalue perturbation analysis:
\begin{equation}\label{formula_eig_perturb1}
\left( {L_P + \delta L_P} \right)\left( {{u_i} + \delta {u_i}} \right) = \left( {{\lambda _i} + \delta {\lambda _i}} \right)\left( {{u_i} + \delta {u_i}} \right),
\end{equation}

where a perturbation $\delta L_P$ that includes a new edge connection  is applied to $L_P$, resulting in perturbed eigenvalues and eigenvectors  ${\lambda _i} + \delta {\lambda _i}$ and ${u_i} + \delta {u_i}$ for $i=1,...,n$, respectively. 

Keeping only the first-order terms leads to:
\begin{equation}\label{formula_eig_perturb1_first_order}
 {L_P}\delta {u_i} + {\delta L_P}{u_i} = {{\lambda _i}{\delta {u_i}} + \delta {\lambda _i}}  {{u_i} }.
\end{equation}
Write $\delta u_i$ in terms of the original eigenvectors $u_i$ for for $i=1,...,n$:
\begin{equation}\label{delta u_i}
{\delta {u_i}} = \sum\limits_{i = 1}^{n} {{\alpha _i}{u_i}}.
\end{equation}

Substituting (\ref{delta u_i}) into (\ref{formula_eig_perturb1_first_order}) leads to:

\begin{equation}\label{formula_eig_perturb1_first_order_expand}
 {L_P}\sum\limits_{i = 1}^{n} {{\alpha _i}{u_i}} + {\delta L_P}{u_i} = {{\lambda _i}\sum\limits_{i = 1}^{n} {{\alpha _i}{u_i}} + \delta {\lambda _i}}  {{u_i} }.
\end{equation}
Multiplying ${u_i^T}$ to both sides of (\ref{formula_eig_perturb1_first_order_expand}) results in:

\begin{equation}\label{formula_eig_perturb1_first_order_multiply}
 {u_i^T}{L_P}\sum\limits_{i = 1}^{n} {{\alpha _i}{u_i}} + {u_i^T}{\delta L_P}{u_i} = {{\lambda _i}{u_i^T}\sum\limits_{i = 1}^{n} {{\alpha _i}{u_i}} + \delta {\lambda _i}}{u_i^T}{{u_i} }.
\end{equation}

Since $u_i$ for for $i=1,...,n$ are unit-length, mutually-orthogonal eigenvectors, we have:

\begin{equation}\label{formula_eig_perturb1_first_order_huajian}
 {u_i^T}{L_P}\sum\limits_{i = 1}^{n} {{\alpha _i}{u_i}}  = {\alpha _i}{u_i^T}{L_P}{{u_i}},~~~~~~~~ {\lambda _i}{u_i^T}\sum\limits_{i = 1}^{n}{\alpha _i}{u_i}={\alpha _i}{u_i^T}{\lambda _i}{u_i}.
\end{equation}

Substituting (\ref{formula_eig_perturb0}) into (\ref{formula_eig_perturb1_first_order_huajian}), we have:

\begin{equation}\label{interium}
 {\alpha _i}{u_i^T}{L_P}{{u_i}}  = {\alpha _i}{u_i^T}{\lambda _i}{u_i}.
\end{equation}

According to (\ref{formula_eig_perturb1_first_order_huajian}), we have:

\begin{equation}\label{formula_eig_perturb1_first_order_sparresult}
 {u_i^T}{L_P}\sum\limits_{i = 1}^{n} {{\alpha _i}{u_i}}  = {\lambda _i}{u_i^T}\sum\limits_{i = 1}^{n}{\alpha _i}{u_i}.
\end{equation}

Substituting (\ref{formula_eig_perturb1_first_order_sparresult}) into (\ref{formula_eig_perturb1_first_order_multiply}) leads to:

\begin{equation}\label{formula_eig_perturb1_first_order_final}
{u_i^T}{\delta L_P}{u_i} = {  \delta {\lambda _i}}{u_i^T}{{u_i} }= \delta {\lambda _i}.
\end{equation}

 Then the eigenvalue perturbation due to ${\delta L_P}$   is given by:

\begin{equation}\label{formula_eig_perturb1_conclusion}
 \delta {\lambda _i} = w_{p,q}\left( {{{u_i^T}e_{p,q}} } \right)^2.
\end{equation}

If each edge weight $w_{p,q}$ encodes the similarity of data vectors $x_p$ and $x_q$ at nodes $p$ and $q$, it can be shown that $w_{p,q} \propto \frac{1}{z^{data}}$, where $z^{data}$ denotes the distance between $x_p$ and $x_q$; on the other hand, $\left( {{{u_i^T}e_{p,q}}  } \right)^2 \propto z^{emb}$. Therefore, as long as we can find an edge with large $w_{p,q}\left( {{{u_i^T}e_{p,q}}  } \right)^2$ or ${\eta_{p,q}}= \frac{z^{emb}_{p,q}}{z^{data}_{p,q}}$, including this edge into the current graph will significantly perturb the Laplacian eigenvalue $\lambda_i$ and eigenvector $u_i$.

\section{Algorithm Flow}

\begin{algorithm}[!htbp]
 { \caption{GRASPEL with Fiedler-vector based spectrally-critical edge identification} \label{alg:graspel}
\noindent\textbf{Input:} A data set with $N$ data points $x_1,...x_N \in \mathbb{R}^{d}$, window size $\epsilon$, edge selection ratio $\zeta$.
\noindent\textbf{Output:} The spectrally-learned graph.\\

\begin{algorithmic}[1]
    \STATE Construct an initial ANN graph $G_i$ as in     \citep{chen2011parallel}. \\
    \STATE Initialize: $Terminate$=0;
     \WHILE{$Terminate$==0}
     \STATE{Embed $G_i$ with the Fiedler vector  and sort the data points (nodes);}
     \STATE{Evaluate the embedding distortions of candidate edges connecting the top  and bottom $\epsilon N$ sorted nodes; }
     \STATE{Select top $\zeta N$   edges based on the evaluation result and add them to $G_i$; }
     \STATE Check the spectral stability and update $Terminate$.
     \ENDWHILE
    
\end{algorithmic}
}
\end{algorithm}

\begin{algorithm}[!htbp]
\small { \caption{Spectral Clustering Algorithm} \label{alg:sc}
\textbf{Input:} A graph $G=(V,E,w)$ and the number of clusters k.\\
\textbf{Output:} Clusters $C_1$...$C_k$.\\

\begin{algorithmic}[1]
    \STATE Compute the adjacency matrix $A_G$, and diagonal matrix $D_G$; \\
    \STATE Obtain the unnormalized Laplacian matrix $L_G$=$D_G$-$A_G$;\\
    \STATE Compute the eigenvectors $u_1$,...$u_k$ that correspond to the bottom k nonzero eigenvalues of $L_G$;\\
    \STATE Construct $U \in \mathbb{R}^{n \times k}$, with k eigenvectors of $L_G$ stored as columns;\\
    \STATE Perform k-means algorithm to partition the rows of $U$ into k clusters and return the result.\\
\end{algorithmic}
}
\end{algorithm}

\section{Data Sets Description}\label{dataApp}

COIL20: A data set contains $1,440$ gray-scale images of 20 objects, and each object on a turntable has 72 normalized gray-scale images taken from different degrees. The image size is 32x 32 pixels.

PenDigits: A data set consists of 7,494 images of handwritten digits from 44 writers, using the sampled coordination information. Each digit is represented by 16 attributes.

USPS: A data set includes $9,298$ scanned hand-written digits on the envelops from U.S. Postal Service with $256$ attributes.

MNIST: A data set consists of 70,000 images of handwritten digits. Each image has 28-by-28 pixels in size. This database can be found from Prof.Yann LeCun's website (http://yann.lecun.com/exdb/mnist/).

\section{Compared Algorithms}
\textbf{Standard kNN}: the most widely used affinity graph construction method. Each node is connected to its $k$ nearest neighbors.

\textbf{Consensus of kNN (cons-kNN) }\citep{premachandran2013consensus}: adopts the state-of-the-art neighborhood selection methods to construct the affinity graphs. It selects strong neighborhoods to improve the robustness of the graph by using the consensus information from different neighborhoods in a given kNN graph. 


{  \textbf{LSGL} \citep{kalofolias2017large}: a method to automatically select the parameters of the
model introduced in \citep{kalofolias2016learn} given a desired graph sparsity level. }

\textbf{GL-SigRep} \citep{dong2016learning}: construct a graph from signals that are assumed to be smooth with respect to the corresponding graph.

\textbf{GL-LogDet} \citep{dong2016learning}: encodes the information about the partial correlations between the variables without the constraint to form a valid Laplacian.

\textbf{Graph Learning under structural constraints (GLSC)} \citep{egilmez2017graph}: formulated the problem as to maximum a posterior estimation of Gaussian Markov Random Field (GMRF) when the precision matrix is chosen to be a graph Laplacian.

\section{Evaluation Metric}
\textbf{Evaluation Metric for Spectral Clustering.} (1) {The ACC metric} measures the agreement between the clustering results generated by clustering algorithms and the ground-truth labels. A higher value  of $ACC$ indicates better clustering quality. The ACC can be computed by:
\begin{equation}\label{eqn:scale}
ACC= \frac{\sum\limits_{j = 1}^n  {\delta {(y_i,map(c_i))}}}{{n}},
\end{equation}
where $n$ is the number of samples in the data set, $y_i$ is the ground-truth label provided by the data sets, and $c_i$ is clustering result obtained from the algorithm. $\delta (x,y)$ is a delta function defined as: $\delta (x,y)$=1 for $x=y$, and $\delta (x,y)$=0,  otherwise. $map(\bullet)$ is a permutation function that maps each cluster index $c_i$  to a ground truth label, which can be realized using the Hungarian algorithm    \citep{papadimitrou1982combinatorial}. 
(2) {The NMI metric} is in the range of [0, 1], while a higher NMI value indicates a better matching between the algorithm generated result and ground truth result.
For two random variables $P$ and $Q$, normalized mutual information is defined as     \citep{strehl2002cluster}:
\begin{equation}\label{eqn:scale2}
NMI= \frac{I(P,Q)}{{\sqrt{H(P)H(Q)}}},
\end{equation}
where $I(P,Q)$ denotes the mutual information between $P$ and $Q$, while $H(P)$ and $H(Q)$ are entropies of $P$ and $Q$. In practice, the NMI metric can be calculated as follows     \citep{strehl2002cluster}:
\begin{equation}\label{eqn:scale3}
NMI= \frac{{\sum\limits_{i = 1}^k}{\sum\limits_{j = 1}^k}{n_{i,j}}\log(\frac{{n}\cdot{n_{i,j}}}{{n_i}\cdot{n_j}}) }{{\sqrt{(\sum\limits_{i = 1}^k {{n _{i}}{\log\frac{n_i}{n}}})(\sum\limits_{j = 1}^k {{n _{j}}{\log\frac{n_j}{n}}})}}},
\end{equation}
where $n$ is the number of  data points in the data set, k is the number of clusters, $n_i$ is the number of data points in cluster $C_i$ according to the clustering result generated by algorithm, $n_j$ is the number of data points in class $C_j$ according to the ground truth labels provided by the data set, and $n_{i,j}$ is the number of data points in cluster $C_i$ according to the clustering result as well as in class $C_j$ according to the ground truth labels. 

\textbf{Evaluation Metric for Graph Recovery.} Four widely adopted evaluation metrics in information retrieval have been adopted: Precision, Recall, F-measure and Normalized Mutual Information (NMI) \citep{dong2016learning}. The Precision measures the percentage of correct edges (the edges that are present in the ground-truth graph) in the learned graph. The Recall measures the percentage of the edges in the ground-truth graph that are also in the learned graph. F-measure measures the overall quality by taking both Precision and Recall into account. NMI measures the mutual dependence between the edge sets of the learned graph and the ground-truth graph. 

\section{Additional Results for GR and DR Applications} 

\subsection{Graph Learning for Graph Recovery (GR)}\label{sec:GR}
\begin{table*}[htb]
\begin{center}
\addtolength{\tabcolsep}{-2.5pt}\centering
\caption{Graph Recovery Results}  
\scalebox{1}{
\begin{tabular}{ |c|c|c|c|c|c|c|c|c|c|c|c}
\hline
 &\multicolumn{4}{|c|}{The Gaussian graph} &\multicolumn{4}{|c|}{The ER graph}\\
 \hline Algorithm&F-measure& Precision & Recall &NMI&F-measure& Precision & Recall &NMI\\
 \hline  GL-SigRep &\textbf{0.8310}&0.8120 &0.8826 &\textbf{0.5272} &0.7243&0.6912&0.8389&0.3600 \\
 \hline  GL-LogDet   &0.8178 & {0.8193}&0.8521&0.4701&\textbf{0.7378}&0.6983&0.8030&\textbf{0.4012}\\
 \hline GLSC   &0.7203&0.6901&0.9000&0.3208&0.6609&0.5427&0.8224&0.3379\\    
 \hline GRASPEL  &\textbf{0.8499}  & {0.8394} &0.8812 &\textbf{0.5397} &\textbf{0.7256}&0.6990&0.8132&\textbf{0.3607}\\
 
 \hline
\end{tabular}}\label{table:rocovery}
\end{center}
\end{table*}
\textbf{GR Results.} We also quantitatively compare graph recovery performance of GRASPEL   with state-of-the-art GSP-based graph learning methods, by comparing the graphs learned from observations to the ground truth. { The experiments are repeated $20$ times  for each of the two widely-used synthetic graphs: \textbf{1)} The Gaussian graph:  the coordinates of the vertices are generated uniformly in the unit square randomly. Edge weights are determined by the Gaussian radial basis function. \textbf{2)} The ER graph: the graphs generated by following the Erdos-Renyi model \citep{erdHos1960evolution}}.
 The   best two F-measure and NMI results have been highlighted in Table~\ref{table:rocovery}, showing the effectiveness of GRASPEL in learning graphs that are always very  close (similar) to the ground-truth graphs. Compared with other graph learning methods that can only deal with a few hundreds or thousands of data entities, GRASPEL shows much better (nearly-linear runtime and space) scalability and thus will be more efficient for handling large data sets. 


\begin{wrapfigure}{R}{2.95025002in}
 \vspace*{-20pt}
 \hspace*{-5 pt}\includegraphics[width=2.95in]{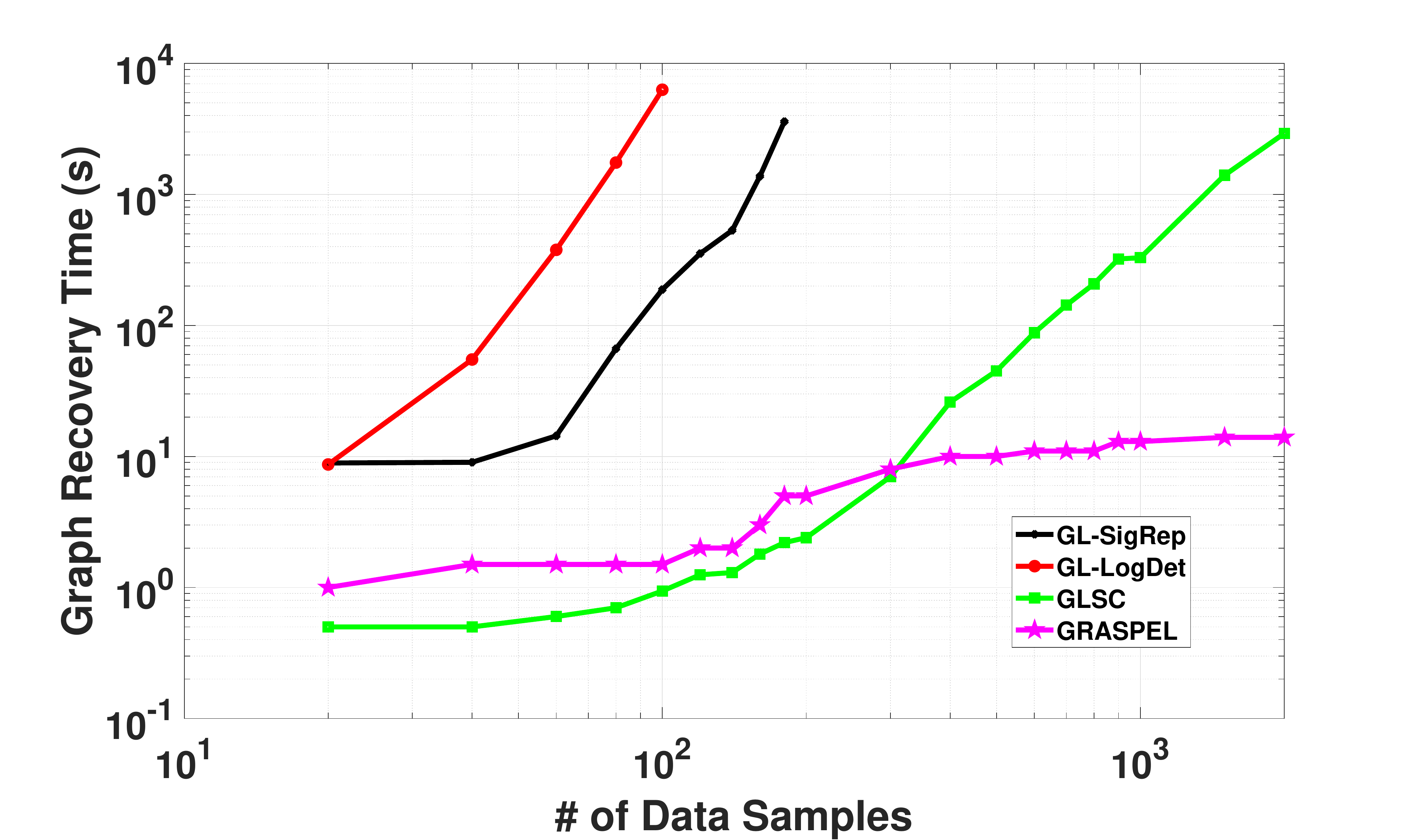}
 \vspace*{-5pt}
 \caption{\small Graph recovery time comparisons.\protect\label{fig:scalabilityNew}}
 \vspace*{0pt}
 \end{wrapfigure}


\textbf{Runtime Scalability.} {As shown in Figure \ref{fig:scalabilityNew}, the graph recovery runtime results of GRASPEL has been compared with   state-of-the-art graph learning methods, such as the GL-SigRep, GL-LogDet and GLSC algorithm proposed in \citep{dong2016learning, egilmez2017graph}. Since the graph learning method LSGL proposed in \citep{kalofolias2017large} can not produce comparable quality of graph recovery results, we did not show the runtime results in the figure. As observed, the proposed approach has a much better runtime scalability when comparing with state-of-the-art methods.}


 \subsection{Graph Learning for Dimensionality Reduction (DR)}\label{sec:DR}

 The t-Distributed Stochastic Neighbor Embedding (t-SNE) has become one of the most popular visualization tools for  high-dimensional data analytic tasks     \citep{maaten2008visualizing,linderman2017clustering}. However, its high computational cost   limits its applicability to large scale problems. {An substantially  improved t-SNE algorithm has been introduced based on tree approximation \citep{van2014accelerating}. However, for large data set the computational cost can still be very high.} 
 
\begin{figure}\centering
\includegraphics[width=0.755\textwidth]{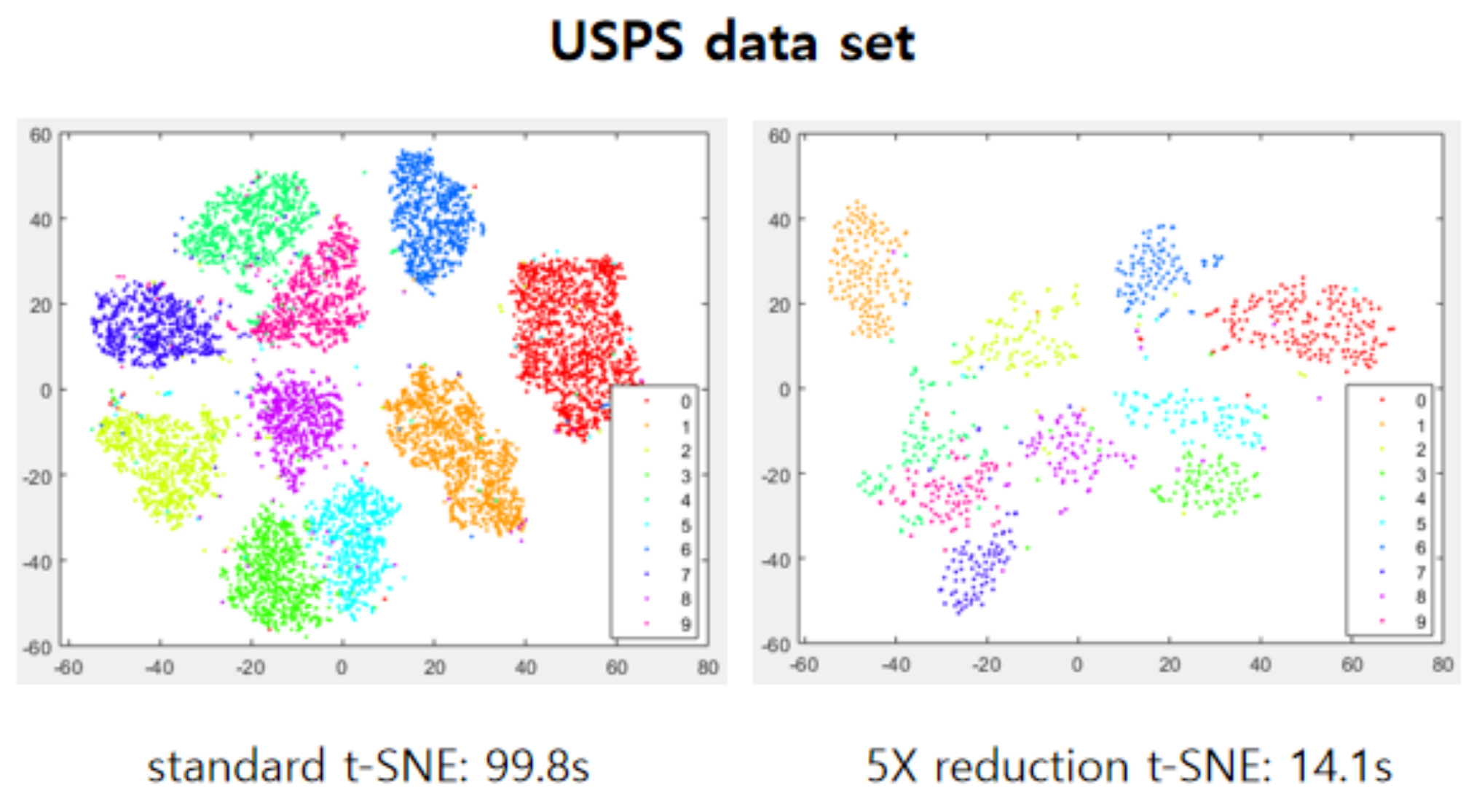}\\
\includegraphics[width=0.755\textwidth]{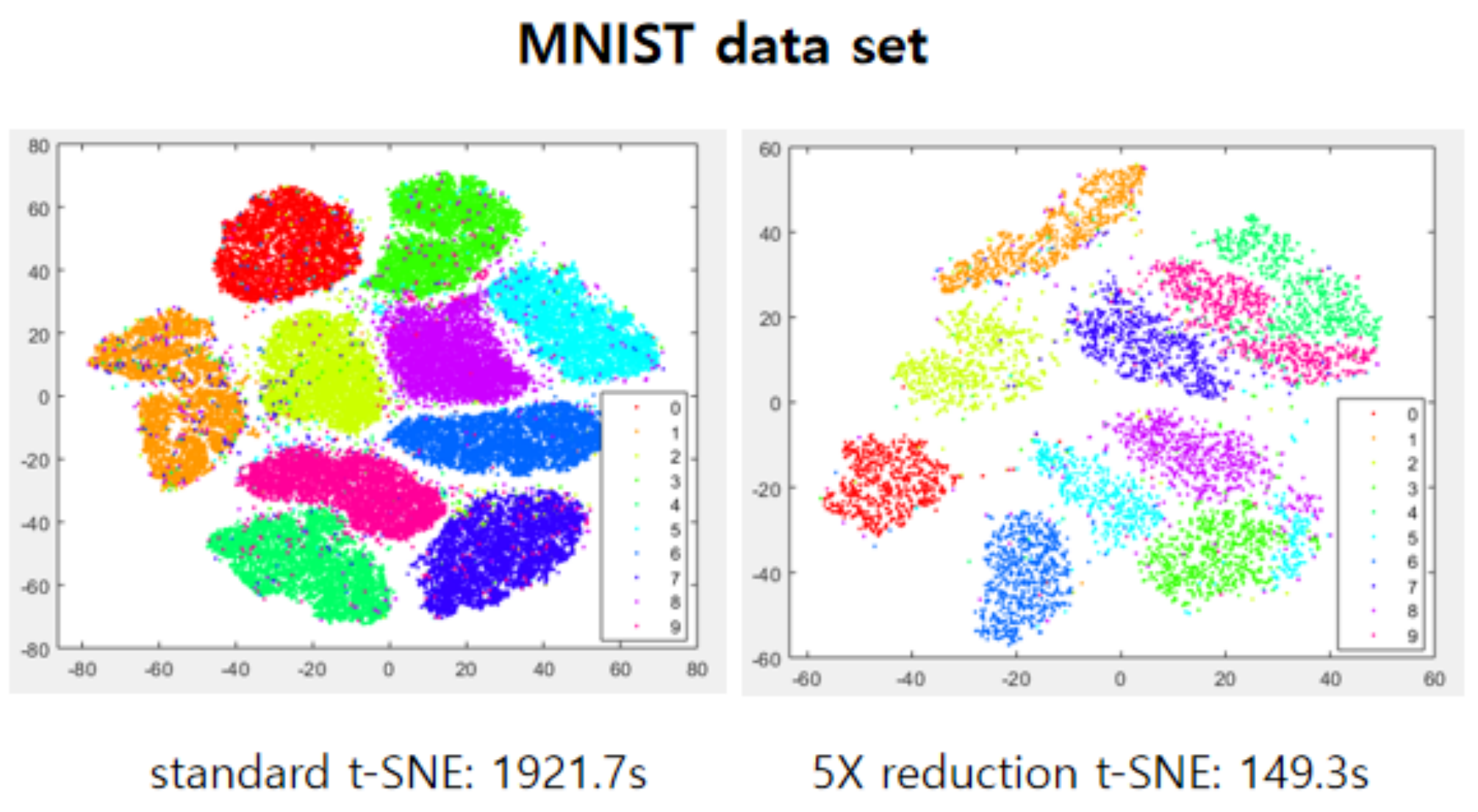}
\caption{t-SNE visualization results.\protect\label{fig:usps5X}}
\end{figure}

 {A multilevel t-SNE algorithm has been proposed in \citep{zhao2018nearly}  leveraging spectral graph coarsening as a pre-processing step applied to  the original kNN graph.  A much smaller set of representative data points can be then selected from the coarsened graph for t-SNE visualization.  In this work, we use GRASPEL to learn  ultra-sparse graphs that can be further reduced into  much smaller ones using spectral graph  reduction  \citep{zhao2018nearly}. Then more efficient t-SNE visualization can be achieved based on the sampled data points corresponding to the nodes in the coarsened graphs. }  
Figure ~\ref{fig:usps5X}     shows the visualization and runtime results of the standard t-SNE (with tree-based acceleration)  \citep{van2014accelerating} and the multilevel t-SNE algorithm \citep{zhao2018nearly} based on   graphs learned by GRASPEL. When using a $5X$ graph reduction ratio,     t-SNE  can be dramatically accelerated (12.8X and 7X speedups for MNIST and USPS data sets, respectively) without loss of visualization quality.

\end{appendices}
\end{document}